\documentclass[letterpaper, 10 pt, conference]{ieeeconf}  
\IEEEoverridecommandlockouts                              
\usepackage{graphicx}
\usepackage{booktabs}
\usepackage{multirow}
\usepackage{cuted}
\usepackage{capt-of}
\usepackage{amsmath}
\usepackage{amsfonts}
\usepackage{tabularx}
\usepackage{array}
\usepackage{float}
\usepackage{placeins}
\usepackage{amsmath}
\usepackage{amssymb}
\usepackage{algorithm}
\usepackage{algpseudocode}
\usepackage{multirow}
\usepackage{bm}
\usepackage{hyperref}
\hypersetup{hidelinks}

\title{\LARGE \bf
Contact Trajectory Prompting: In-Context Transfer of Contact-Rich Behaviors from a Single Demonstration
}

\author{Xian Nie\textsuperscript{1,2*}, Yujie Zang\textsuperscript{1,2,3*}, Yuhang Zheng\textsuperscript{3*}, Yupeng Zheng\textsuperscript{4}, Songen Gu\textsuperscript{5}, \\ 
Wendi Chen\textsuperscript{1}, Chuan Wen\textsuperscript{1}, Cewu Lu\textsuperscript{1}, Wenchao Ding\textsuperscript{2}, Junchi Yan\textsuperscript{1\textdagger}, Shuicheng Yan\textsuperscript{3\textdagger}%
\\
{\small \textsuperscript{1}Shanghai Jiao Tong University \quad \textsuperscript{2}TARS Robotics \quad \textsuperscript{3}National University of Singapore}\\
{\small \textsuperscript{4}Institute of Automation, Chinese Academy of Sciences \quad \textsuperscript{5}Fudan University}\\
{\small \textsuperscript{*}Equal contribution \quad \textsuperscript{\textdagger}Corresponding authors}
}

\begin{document}

\maketitle
\thispagestyle{empty}
\pagestyle{empty}

\begin{strip}
    \centering
    \includegraphics[width=\textwidth]{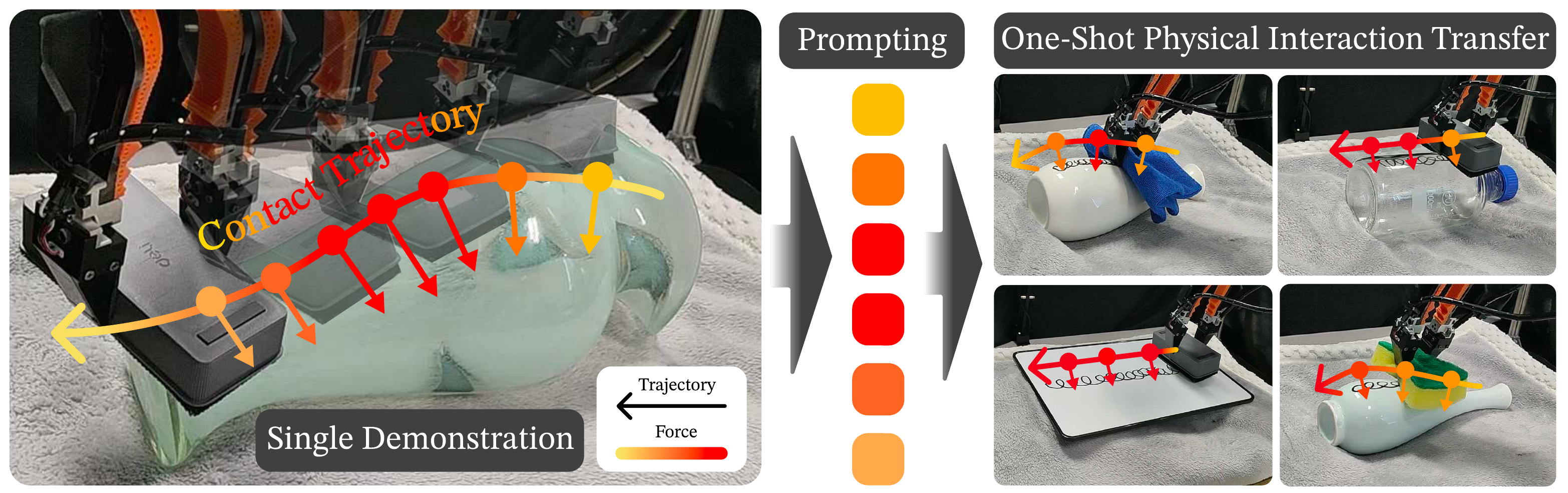}
    \vspace{-3mm}
    \captionof{figure}{
    \textbf{Contact Trajectory Prompting (CTP).}
    CTP encodes motion and contact measurements from a single demonstration into an ordered prompt that captures their temporal evolution. The prompt specifies the desired interaction behavior, including contact intensity and its temporal evolution, while online observations guide a frozen policy in closed-loop execution on unseen object instances. This allows the demonstration to convey how the robot should physically interact with an object, alongside the demonstrated motion.
    }
    \label{fig:teaser}
\end{strip}
\vspace{-5mm}

\begin{abstract}
In-context robot learning enables robots to acquire new behaviors from demonstrations without test-time parameter updates.
However, visual and kinematic specifications alone may not distinguish contact-rich behaviors that share similar motions but require different interaction intensities or patterns.
We propose Contact Trajectory Prompting (CTP), a framework for one-shot transfer of contact-rich manipulation behaviors.
CTP encodes motion and contact information from a single demonstration into an ordered contact trajectory prompt that captures their temporal evolution and specifies the desired physical interaction behavior. 
During execution, the policy uses online sensory feedback to retrieve interaction-relevant information from the prompt, conditioning closed-loop action generation on the demonstrated contact behavior.
Across real-world wiping, rotation, and insertion tasks on objects excluded from training, CTP achieves 78.7\% overall success, outperforming the strongest evaluated baseline by 13.0 percentage points.
Changing only the reference demonstration enables CTP to follow contact specifications for both intensity and direction, including an intermediate wiping force level and clockwise or counterclockwise cap rotation.
These results demonstrate that contact trajectories serve as effective behavioral specifications, extending in-context robot learning beyond motion imitation to physical interaction transfer.
Project website: \url{https://ctp-robotics.github.io/project_page/}.

\end{abstract}

\section{INTRODUCTION}

Learning new manipulation behaviors from few or even single demonstrations is a fundamental challenge in robot learning. Recent advances in in-context robot learning (ICRL) provide a promising framework by enabling robots to adapt their behavior at test time through conditioning on behavior demonstrations, without task-specific parameter updates~\cite{zhou2026zero, jiang2026robottt,wang2026context}. 
In this paradigm, demonstrations serve as contextual specifications that convey task objectives and execution patterns, allowing the policy to infer the desired behavior and adapt it to current observations~\cite{nguyen2026iclr}.

Existing ICRL methods use visual observations~\cite{jain2024vid2robot}, language instructions~\cite{yin2025context}, or action trajectories~\cite{patel2026behavior} to specify task objectives and execution behaviors, enabling robots to adapt their actions to the provided context. In contact-rich tasks, the desired behavior also involves how strongly the robot should interact with an object, which may not be fully specified by visual and kinematic cues alone.
For example, under the same wiping task configuration, a robot may follow similar motion paths, while light and heavy wiping require different contact intensities.
Thus, demonstrations with similar motions may convey different interaction requirements, and reproducing their motion alone may not achieve the desired interaction intensity.

Representing these interaction requirements calls for encoding contact intensity and its temporal evolution. Force and tactile sensing provide direct access to this information. However, related studies primarily use contact information for feedback control or training supervision~\cite{lou2026dream, ye2025learning, zheng2026omnivta}, with less emphasis on using reference contact to specify the desired interaction intensity and its temporal evolution. This motivates us to investigate how demonstrated contact patterns can serve as behavioral specifications, guiding a policy to follow the reference intensity while adapting to the current contact state through online feedback.


To this end, we introduce Contact Trajectory Prompting (CTP), a framework for transferring contact-rich behaviors from a single demonstration. CTP encodes demonstrated motion and contact measurements into an ordered prompt that preserves their temporal evolution and specifies the desired interaction behavior. During execution, a frozen policy uses online observations to retrieve relevant information from this prompt and generate actions that follow the reference specification while adapting to the current contact state.

We evaluate CTP on three real-world contact-rich manipulation task families: wiping, rotation, and insertion. Across object-disjoint one-shot transfer experiments, CTP enables robots to transfer demonstrated contact-rich behaviors from a single reference demonstration without policy updates. Furthermore, changing only the reference demonstration allows the robot to reproduce distinct interaction behaviors, including different contact intensities and manipulation directions. These results demonstrate that contact trajectories can serve as expressive behavioral specifications, extending in-context robot learning beyond motion imitation toward physical interaction strategy transfer.

The main contributions of this paper are as follows:
\begin{itemize}
    \item We formulate reference contact trajectories as behavioral specifications for one-shot manipulation within known task families.

    \item We develop a contact prompt encoder that compresses demonstrated motion and bilateral contact measurements into ordered tokens, together with observation-conditioned prompt retrieval for action generation.

    \item We evaluate the approach on three real-world task families and analyze the roles of reference contact, online contact feedback, and prompt representation.
\end{itemize}
\section{RELATED WORKS}

\subsection{In-Context Robot Learning}
Recent robot learning methods increasingly use demonstrations not only as training supervision, but also as task specifications at inference time~\cite{sridhar2025ricl}.
Vid2Robot~\cite{jain2024vid2robot} conditions robot policies on human demonstration videos to transfer visuomotor behaviors, while Keypoint Action Tokens (KAT)~\cite{di2024keypoint} cast few-shot imitation as prediction over structured visual and action tokens. Within this paradigm, in-context imitation learning directly infers new behaviors from demonstrations provided in context. In-Context Robot Transformer (ICRT)~\cite{fu2025icrt} formulates this process as next-token prediction over sensorimotor trajectories, whereas Instant Policy~\cite{vosylius2025instant} uses graph diffusion to infer manipulation actions from a few demonstrations.
Beyond task specification, several methods explore how demonstration context can encode and control specific behaviors. Behavior Prompting Policy (BPP)~\cite{patel2026behavior} explicitly treats a single sensorimotor demonstration as a behavior prompt, enabling a fixed policy to adapt its behavior at inference time. Instant-Fold~\cite{wang2026instant} similarly conditions a policy on a single demonstration to specify different manipulation modes for deformable objects.

These works highlight the use of demonstrations to specify robot behaviors at inference time. While these methods emphasize visual and kinematic information~\cite{nguyen2026iclr,sridhar2025ricl,Dreczkowski2025LearningAT}, our approach incorporates the temporal evolution of contact forces into demonstration context to specify the desired physical interaction behavior.

\subsection{Contact-aware Manipulation}
Force and tactile sensing have been increasingly incorporated into robot learning to capture interaction dynamics and improve control in contact-rich manipulation~\cite{hou2025adaptive}.
Recent works use contact signals as online feedback for responsive closed-loop control, including Force Policy~\cite{fang2026force}, Reactive Diffusion Policy (RDP)~\cite{xue2025reactive}, and AT-VLA~\cite{li2026vla}. 
Beyond reactive feedback, recent methods predict future force or tactile signals to anticipate interaction dynamics. FoAR~\cite{he2025foar}, $N_0$-TWAM~\cite{team2026n_0}, OmniVTA~\cite{zheng2026omnivta}, and related approaches~\cite{zang2026tacforesight,tian2026vt,yuan2026ftp,ma2026tacpac} model the temporal evolution of contact to support contact-aware action generation.
Together, these works highlight the value of modeling contact signals over time for contact-aware manipulation.

Another direction explores physical interaction information as temporal context or higher-level conditioning for contact-aware behavior. 
LIFT~\cite{wang2026never} incorporates recent force observations through causal force memory, while FM-VLA~\cite{li2026fm} compresses accumulated force histories into memory tokens to reason about contact events and interaction progress. ForceVLA2~\cite{li2026forcevla2} further introduces force-based prompts to construct force-aware task concepts for manipulation. 
Our focus is the contact information contained in a reference demonstration: it specifies the desired interaction behavior, whereas online contact measurements describe the robot’s current interaction. CTP encodes the former into an ordered prompt and uses the latter, together with visual and proprioceptive observations, to condition execution.

\begin{figure*}[t]
    \centering
    \includegraphics[width=\textwidth]{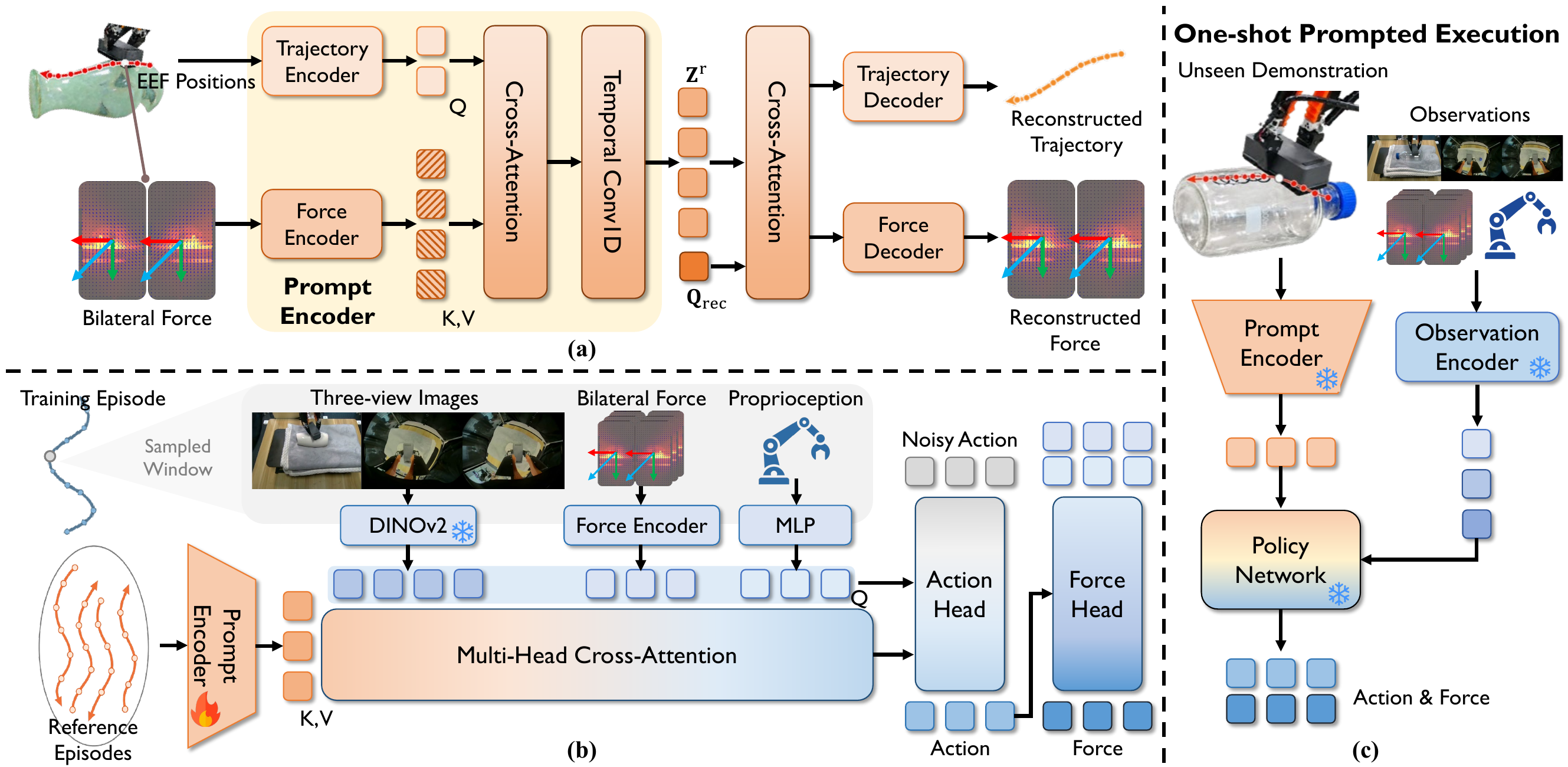}
    \vspace{-4mm}
    \caption{
    \textbf{Overview of CTP.}
    (a) \textbf{Prompt pretraining:} The prompt encoder maps motion and force signals into ordered tokens and is pretrained through reconstruction.
    (b) \textbf{Prompt-conditioned policy training:}
    Online observations attend to a reference prompt from a different demonstration of the same behavior, conditioning action generation with auxiliary force supervision.
    (c) \textbf{One-shot execution:} A single demonstration enables in-context adaptation of contact-rich behaviors with a frozen policy.
    }
    \label{fig:pipeline}
    \vspace{-3mm}
\end{figure*}
\section{Method}
\label{sec:method}

We propose \textbf{Contact Trajectory Prompting} for in-context learning of contact-rich behaviors from a single demonstration. As shown in Fig.~\ref{fig:pipeline}, CTP encodes demonstrated motion, distributed force maps, and resultant wrenches into an ordered contact prompt (Sec.~\ref{sec:force_prompt}), which guides a closed-loop policy conditioned on online visual and force observations (Sec.~\ref{sec:policy}). At test time, a new demonstration is encoded once and reused throughout execution, with both the prompt encoder and the policy network kept frozen (Sec.~\ref{sec:one_shot_protocol}).


\subsection{Contact Trajectory Prompt Encoding}
\label{sec:force_prompt}

The prompt encoder maps a reference demonstration containing distributed force maps and resultant 6D wrenches from both fingers into a fixed-length sequence of interaction tokens. These tokens encode the temporal evolution of contact forces and torques during manipulation and serve as a prompt to guide policy execution.

\subsubsection{Contact-aware Reference Representation}

We represent each reference demonstration using end-effector positions, fingertip wrenches, and distributed force maps. These quantities describe the demonstrated motion, the resultant contact wrenches, and the spatial distribution of contact forces, respectively. A reference demonstration is defined as

\[
\mathcal{D}^{r}
=
\{\mathbf{x}^{r}_{t}\}_{t=1}^{T_r},
\qquad
\mathbf{x}^{r}_{t}
=
\bigl(
\mathbf{p}^{r}_{t},
\mathbf{w}^{r}_{L,t},
\mathbf{w}^{r}_{R,t},
\mathbf{F}^{r}_{L,t},
\mathbf{F}^{r}_{R,t}
\bigr).
\]
Here, $\mathbf{p}^{r}_{t}\in\mathbb{R}^{3}$ denotes the end-effector position relative to the start of the demonstration; $\mathbf{w}^{r}_{L,t},\mathbf{w}^{r}_{R,t}\in\mathbb{R}^{6}$ denote the resultant forces and torques at the left and right fingertips; and $\mathbf{F}^{r}_{L,t},\mathbf{F}^{r}_{R,t}$ are normalized fingertip displacement maps used as proxies for distributed contact forces.

To accommodate demonstrations of varying lengths, we use contact-aware hybrid sampling to select a fixed number of frames. We first select $S_u$ uniformly spaced temporal anchors to initialize the index set $\mathcal{S}$, then iteratively add $S_w$ frames through farthest-point sampling in wrench space:
\[
t^{\star}
=
\arg\max_{t\notin\mathcal{S}}
\min_{k\in\mathcal{S}}
\left\|
\widetilde{\mathbf{w}}^{r}_{t}
-
\widetilde{\mathbf{w}}^{r}_{k}
\right\|_2^2,
\qquad
\mathcal{S}\leftarrow\mathcal{S}\cup\{t^{\star}\},
\]
where $\widetilde{\mathbf{w}}^{r}_{t}\in\mathbb{R}^{12}$ is the normalized concatenation of the left- and right-fingertip wrenches. 

Uniform sampling provides temporal coverage, while wrench-space sampling retains distinctive force and torque states that uniform sampling may miss.
The selected frames are sorted chronologically, with positions, wrenches, and distributed force maps sampled at the same instants. With $S_u=S_w=64$, this yields a unified reference representation $\mathbf{X}^{r}$ containing $S=128$ temporally ordered frames.

\subsubsection{Prompt Encoder}
We first encode motion and force measurements separately at each sampled frame. A 2D CNN shared across the two fingertips extracts spatial features from the distributed force maps, followed by spatial average pooling to obtain one token per fingertip. In parallel, a shared MLP encodes each fingertip's 6D wrench into a token. Within each fingertip, the force-map and wrench tokens are concatenated along the feature dimension, yielding separate left and right force features. Another MLP encodes the end-effector positions into temporally aligned motion features.
At each sampled frame, cross-attention uses the motion feature as the query and the left and right force features as keys and values to produce a fused motion--force feature.
This keeps the two fingertips distinct during fusion, while spatial pooling summarizes each force map into a single token.

To capture the temporal evolution of motion and contact forces, we encode the fused feature sequence using a dilated temporal convolutional network (TCN)~\cite{bai2018empirical}. Before temporal encoding, we augment each fused feature with a sinusoidal encoding of its normalized timestamp in the original demonstration. This provides the TCN with the original timing information of the nonuniformly sampled frames. Denoting the augmented sequence by $\mathbf{U}^{r}$, we obtain the contact trajectory prompt as
\[
\mathbf{Z}^{r}
=
\operatorname{AvgPool}_{S\rightarrow K}
\bigl(\operatorname{TCN}_{\phi}(\mathbf{U}^{r})\bigr)
\in \mathbb{R}^{K\times d},
\]
where adaptive average pooling aggregates the encoded features into $K=16$ temporally ordered interaction tokens for policy conditioning.

During reconstruction pretraining only, we randomly mask one fingertip's wrench and distributed force map before feeding them into their respective encoders.


\subsubsection{Prompt Decoder}

We pretrain the prompt encoder by reconstructing the sampled reference from $\mathbf{Z}^{r}$. Learnable queries retrieve frame-level features through cross-attention:
\[
\mathbf{H}^{r}_{\mathrm{dec}}
=
\mathbf{Q}_{\mathrm{rec}}
+
\operatorname{MHA}
\bigl(
Q=\mathbf{Q}_{\mathrm{rec}},
K=\mathbf{Z}^{r},
V=\mathbf{Z}^{r}
\bigr).
\]

A trajectory decoder reconstructs end-effector positions, while a force decoder reconstructs bilateral 6D wrenches and distributed force maps. The encoder and decoders are jointly trained with reconstruction losses on these quantities, using the complete sampled reference as the target even when one fingertip's force inputs are masked. This objective encourages the prompt to retain both motion and contact information and capture their dependencies.

After pretraining, we discard the decoders and jointly optimize the prompt encoder and policy.



\subsection{Prompt-based Policy Training}
\label{sec:policy}


We next train a prompt-conditioned policy using conditional flow matching and an auxiliary force-prediction objective. Specifically, each training sample pairs an observation--action window from episode $i$ with a reference demonstration from a different episode $r\neq i$ of the same behavior. The former provides observations and supervision targets, while the latter provides the contact prompt.

\subsubsection{Online Interaction Encoding}

As shown in Fig.~\ref{fig:pipeline}(b), online RGB observations, proprioceptive states, bilateral wrenches, and distributed force maps are encoded into modality-specific tokens. A frozen DINOv2 backbone~\cite{oquab2023dinov2} extracts visual features, which are projected into the token embedding space by a trainable MLP. Separate MLPs encode proprioceptive and wrench histories, while a shared 2D CNN encodes the distributed force maps of both fingertips. We add temporal and modality embeddings, along with left/right embeddings for fingertip tokens, and concatenate the tokens into the online sequence $\mathbf{Q}_{t}^{0}\in\mathbb{R}^{N_q\times d}$.

To retrieve reference information based on the current observations, we project the contact prompt $\mathbf{Z}^{r}$ into the same feature space as the online tokens and add learnable position embeddings, yielding $\mathbf{R}^{r}\in\mathbb{R}^{K\times d}$. A multi-head cross-attention block then uses the online tokens as queries and the prompt tokens as keys and values:
\[
\mathbf{Q}_{t}^{\ell+1}
=
\mathbf{Q}_{t}^{\ell}
+
\operatorname{MHA}_{\ell}
\bigl(
Q=\mathbf{Q}_{t}^{\ell},
K=\mathbf{R}^{r},
V=\mathbf{R}^{r}
\bigr).
\]

This allows the policy to retrieve demonstrated motion and force information relevant to the current interaction. After $L$ layers, the resulting tokens are pooled within each modality and aggregated across modalities to obtain a prompt-conditioned interaction feature $\mathbf{c}_{t}^{\mathrm{int}}\in\mathbb{R}^{256}$. In parallel, a separate observation encoder fuses vision and recent proprioception into $\mathbf{c}_{t}^{\mathrm{obs}}\in\mathbb{R}^{256}$, providing a direct representation of the current state.
The two features are concatenated into a 512-dimensional condition vector $\mathbf{c}_{t}$ for action generation.

\subsubsection{Flow-matching Action Head}

Given $\mathbf{c}_{t}$, the policy generates an $H$-step action chunk $\mathbf{A}_{t}$ by conditional flow matching. With noise $\boldsymbol{\varepsilon}$ and interpolation time $\tau\in[0,1]$, we set $\mathbf{A}_{t,\tau}=(1-\tau)\boldsymbol{\varepsilon}+\tau\mathbf{A}_{t}$ and target velocity $\mathbf{U}_{t,\tau}=\mathbf{A}_{t}-\boldsymbol{\varepsilon}$, and then optimize
\[
\mathcal{L}_{\mathrm{FM}}
=
\mathbb{E}
\Bigl[
\bigl\|
v_{\theta}(\mathbf{A}_{t,\tau},\tau,\mathbf{c}_{t})
-
\mathbf{U}_{t,\tau}
\bigr\|_{F}^{2}
\Bigr].
\]

An auxiliary head predicts short-horizon changes in bilateral fingertip wrenches from the generated action chunk and \textit{online} state and wrench histories, without direct access to the reference prompt:
\[
\Delta\widehat{\mathbf{w}}_{t+1:t+h}
=
G_{\eta}
\bigl(
\widehat{\mathbf{A}}_{t:t+h-1},
\mathbf{s}_{\leq t},
\mathbf{w}_{\leq t}
\bigr),
\]
where $\mathbf{w}_{t}\in\mathbb{R}^{12}$ concatenates
the left and right fingertip wrenches. Supervising $\mathbf{w}_{t}+\Delta\widehat{\mathbf{w}}$ against future wrench measurements encourages the action chunk to be consistent with contact dynamics, rather than decoding the desired force from the demonstration. The policy objective is
\[
\mathcal{L}_{\mathrm{policy}}
=
\mathcal{L}_{\mathrm{FM}}
+
\lambda_{\mathrm{force}}\mathcal{L}_{\mathrm{force}}.
\]

\subsection{One-shot Prompted Execution}
\label{sec:one_shot_protocol}

At test time, the robot receives a single reference demonstration $\mathcal{D}^{r}$ on an unseen object within the same task. The prompt encoder, the online observation encoder, and the policy network remain frozen. The demonstration is encoded once into $\mathbf{Z}^{r}$ and cached throughout execution.

As summarized in Algorithm~\ref{alg:one_shot_transfer}, the policy uses online visual, proprioceptive, and force observations to retrieve relevant information from the cached prompt and generate an action chunk. The robot executes the first portion of the chunk, then updates its observations and generates the next chunk, enabling receding-horizon control.



\begin{algorithm}[h]
\caption{One-shot execution with a cached contact prompt.}
\label{alg:one_shot_transfer}
\begin{algorithmic}[1]
\Require frozen encoder and policy; one demonstration $\mathcal{D}^{r}$ on the target object
\State $\mathbf{Z}^{r}\gets\textsc{EncodePrompt}(\mathcal{D}^{r})$
\Comment{encode once and cache $K$ tokens}
\State Reset the robot
\While{not terminated}
    \State $\mathbf{o}_{t}\gets$ vision, proprioception, wrench, tactile
    \State $\mathbf{c}_{t}\gets\textsc{Retrieve}(\mathbf{o}_{t}\rightarrow\mathbf{Z}^{r})$
    \Comment{online contact queries the cached prompt}
    \State $\mathbf{A}_{t}\gets\textsc{FlowMatch}(\mathbf{c}_{t})$
    \State Execute the first $H/2$ actions of $\mathbf{A}_{t}$
\EndWhile
\end{algorithmic}
\end{algorithm}

\begin{figure*}[!t]
    \centering
    \includegraphics[width=\textwidth]{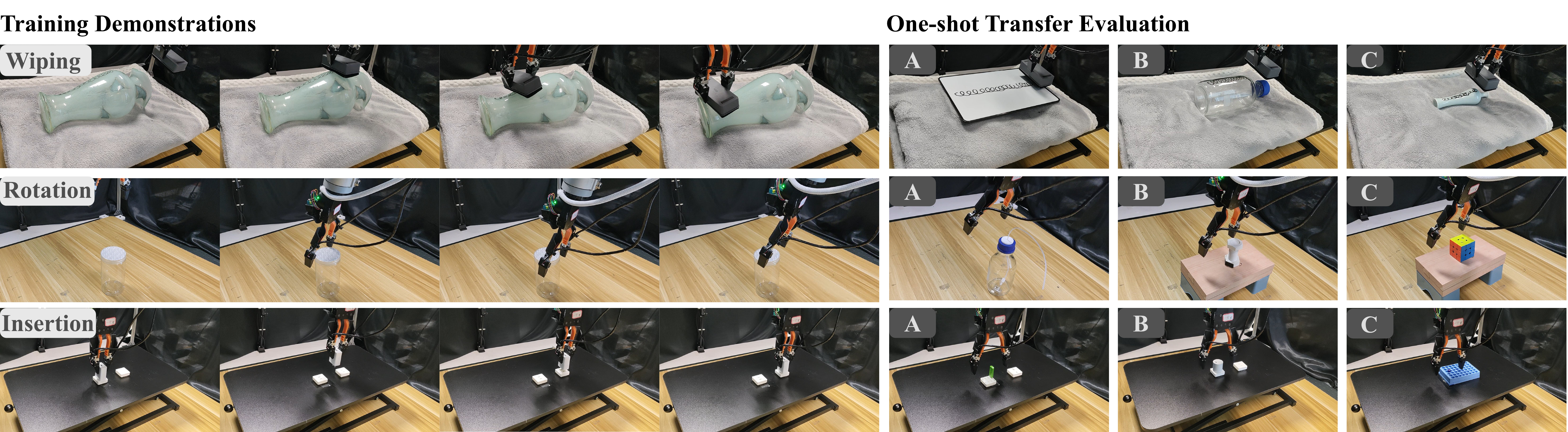}
    \vspace{-5mm}
    \caption{
    \textbf{Benchmark tasks and one-shot transfer evaluation.}
    Training demonstrations (left) specify wiping, rotation, and insertion behaviors.
    The policy is evaluated on unseen objects (right) with similar physical interaction patterns.
    A--C denote the unseen instances used for each task family.
    }
    \vspace{-3mm}
    \label{fig:tasks}
\end{figure*}

\begin{figure}[!t]
    \centering
    \includegraphics[width=\columnwidth]{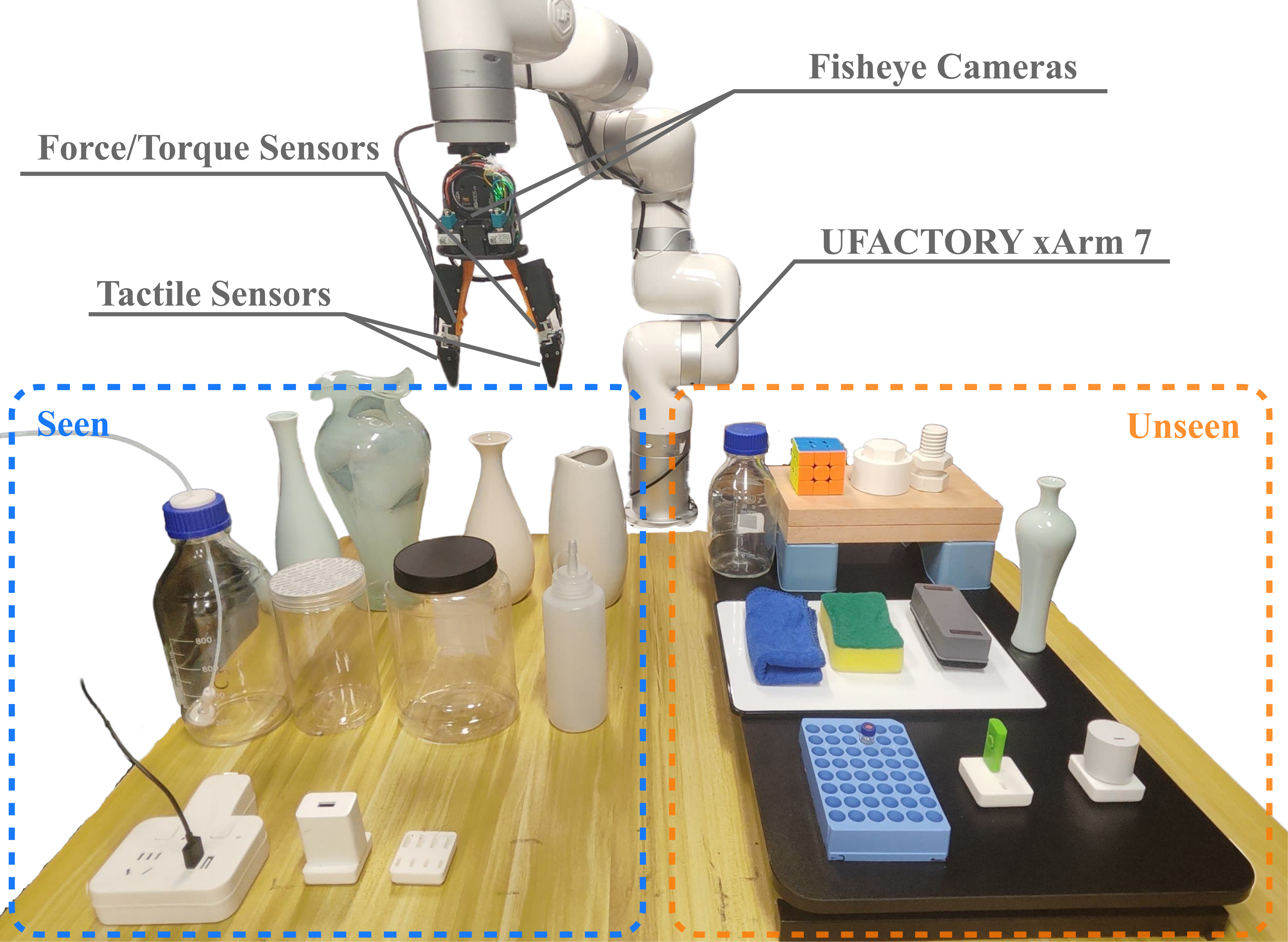}
    \vspace{-2mm}
    \caption{\textbf{Real-world experimental setup for one-shot behavior transfer.} Objects on the left are used for training. Objects on the right are unseen during training and used for one-shot evaluation.}
    \vspace{-5mm}
    \label{fig:setup}
\end{figure}

\section{EXPERIMENTS}
\label{sec:experiments}

Our experiments address three questions:
(1) Can a single demonstration guide a frozen policy on unseen object instances and specify distinct contact behaviors?
(2) What roles do the reference contact trajectory and online contact feedback play in prompted execution?
(3) How do architectural choices affect performance?



\subsection{Experimental Setup}

\subsubsection{Robotic Platform}
Our experimental platform is built upon a 7-DoF UFactory xArm7 robot equipped with a customized gear-driven gripper, as shown in Fig.~\ref{fig:setup}. 
A global Intel RealSense RGB-D camera and two local fisheye cameras mounted on the end-effector provide scene-level and close-range images, respectively. 
Each fingertip carries an Xense optical tactile sensor and a Photonic FingerMAX miniature six-axis F/T sensor to capture local contact deformation and resultant forces and torques, respectively.
Each optical tactile sensor outputs a $35\times20$ map of 3D displacements. We use the normalized displacement maps as proxies for distributed contact forces, hereafter referred to as distributed force maps. All sensory streams are collected and synchronized at 30 Hz.
Demonstrations are collected through PICO-based teleoperation with force-dependent vibrotactile feedback.

\subsubsection{Implementation Details}
During training, the prompt encoder maps each reference trajectory into $K=16$ prompt tokens with a feature dimension of 128. Online query tokens are constructed from the latest RGB observation and an eight-step history of proprioceptive states, bilateral fingertip wrenches, and distributed force maps.
The online grounding module employs three Transformer decoder layers with four attention heads, and the force consistency loss is weighted by $\lambda_{\mathrm{force}}=0.1$.
During inference, each action is represented as a 10-dimensional vector consisting of 3D translation, a 6D continuous rotation representation, and one gripper command. 
The flow-matching policy generates a 32-step relative action chunk from Gaussian noise by integrating the learned velocity field with 16 Euler steps, and the first 16 actions are executed. 
All policy inference is performed on a single NVIDIA GeForce RTX~4090 GPU.

\subsubsection{Baselines}
We compare CTP with some advanced fine-tuning-based and demonstration-prompted baselines. All methods are evaluated on the same robot and objects using the same success criteria.

\begin{itemize}
    \item \textbf{DP-FT}~\cite{chi2025diffusion}: a diffusion policy adapted with bilateral fingertip wrench and distributed force map inputs, \textbf{trained} on the same task demonstrations, and \textbf{fine-tuned} at test time using a single reference trajectory.
    
    \item \textbf{$\bm{\pi_{0.5}}$-FT}~\cite{intelligence2025pi_}: a large-scale pretrained vision--language--action policy adapted with bilateral fingertip wrench and distributed force map inputs, \textbf{post-trained} on the same task demonstrations, and \textbf{fine-tuned}
    at test time using a single reference trajectory.

    \item \textbf{BPP*}~\cite{patel2026behavior}: a behavior prompting policy augmented with bilateral fingertip wrench and distributed force map inputs. It conditions action generation on a reference demonstration represented as visual, proprioceptive, and action tokens, with \textbf{parameters frozen} at test time.
\end{itemize}


\begin{table*}[!t]
\centering
\caption{One-shot transfer to unseen objects. For each task family, A--C match the instances in Fig.~\ref{fig:tasks}.}
\label{tab:main_results}
\vspace{-2mm}
\small
\setlength{\tabcolsep}{8pt}
\renewcommand{\arraystretch}{1.15}

\begin{tabular}{lccc ccc ccc c}
\toprule
\multirow{2}{*}{Method}
& \multicolumn{3}{c}{Wiping} 
& \multicolumn{3}{c}{Rotation}
& \multicolumn{3}{c}{Insertion}
& \multirow{2}{*}{Overall} \\
\cmidrule(lr){2-4}
\cmidrule(lr){5-7}
\cmidrule(lr){8-10}

& A & B & C
& A & B & C
& A & B & C
& \\
\midrule

DP-FT
& 16.7\% & 33.3\% & 41.7\%
& \textbf{100.0\%} & 66.7\% & 0.0\%
& 75.0\% & 0.0\% & 33.3\%
& 40.7\% \\

$\pi_{0.5}$-FT
& 58.3\% & 33.3\% & 8.3\%
& \textbf{100.0\%} & 8.3\% & 8.3\%
& 66.7\% & 50.0\% & 0.0\%
& 37.0\% \\

BPP*
& \textbf{83.3\%} & 41.7\% & 50.0\%
& \textbf{100.0\%} & 91.7\% & 58.3\%
& 66.7\% & 58.3\% & \textbf{41.7\%}
& 65.7\% \\

\textbf{CTP}
& \textbf{83.3\%} & \textbf{83.3\%} & \textbf{66.7\%}
& \textbf{100.0\%} & \textbf{100.0\%} & \textbf{91.7\%}
& \textbf{83.3\%} & \textbf{66.7\%} & 33.3\%
& \textbf{78.7\%} \\

\bottomrule
\vspace{-2mm}
\end{tabular}
\end{table*}

\begin{figure*}[t]
    \centering
    \includegraphics[width=0.95\textwidth,trim=0 6pt 0 0,clip]{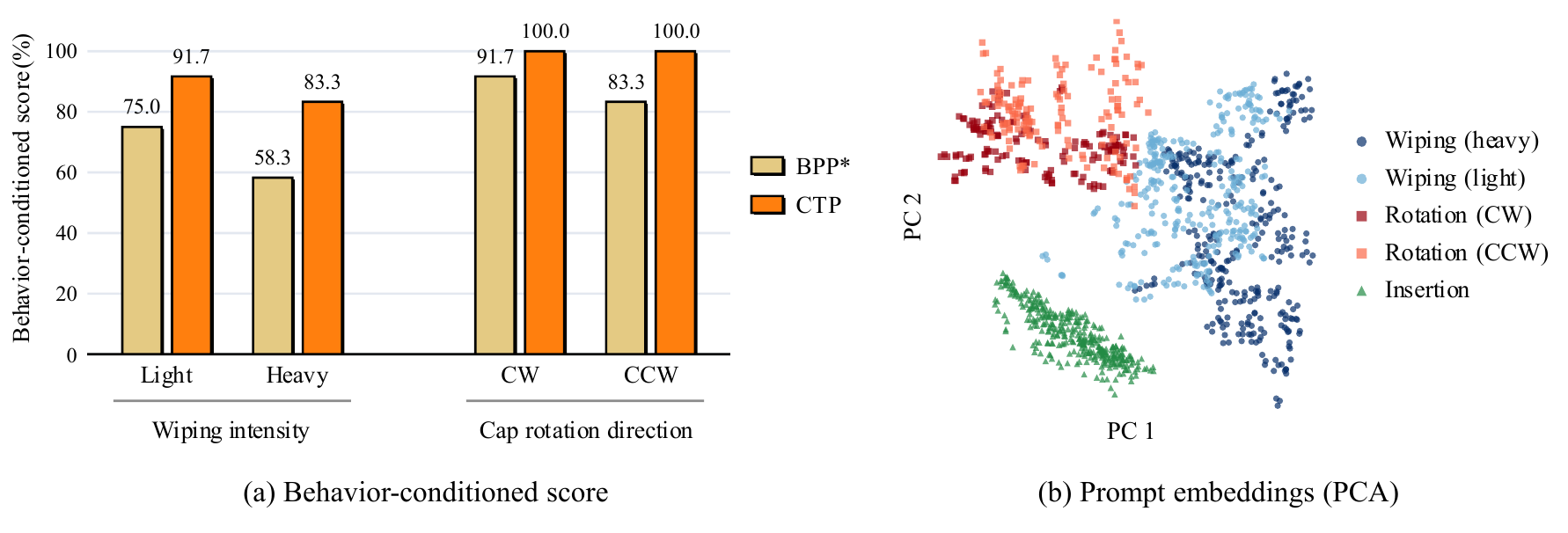}
    \vspace{-3mm}
    \caption{\textbf{Behavior specification with contact prompts.}
    (a)~Behavior-conditioned score when changing only the reference demonstration to specify wiping intensity or cap rotation direction.
    (b)~PCA of contact prompt embeddings, colored by task and behavior.}
    \vspace{-3.5mm}
    \label{fig:specification}
\end{figure*}

\subsection{Evaluation Protocols and Benchmark Tasks}

\subsubsection{Evaluation Protocols}

We evaluate one-shot transfer to objects unseen during policy training. For each trial, the policy receives a single reference demonstration collected on the test object (Fig.~\ref{fig:tasks}). CTP encodes this demonstration once and reuses the prompt throughout the trial, keeping the encoder and policy fixed.


For each method and test object in Table~\ref{tab:main_results}, we evaluate four reference demonstrations with three trials each (12 trials total) and report the task success rate. We also report behavior-conditioned scores for wiping intensity and cap rotation direction: each trial scores 1 if it completes the task and matches the reference behavior, and 0 otherwise. Scores are averaged and reported as percentages.

\subsubsection{Tasks and Behaviors}

\begin{itemize}
    \item \textbf{Wiping}: The robot grasps an eraser and wipes a target region. Success requires covering the region while maintaining stable tool--surface contact. 
    For each reference, we define a force band as the temporal mean fingertip force $\pm$ one standard deviation during the wiping phase. Intensity is considered matched if the executed force falls within this band for at least 50\% of the executed wiping phase. 
    We evaluate on unseen wiping objects and surfaces, using light and heavy references to test control of wiping intensity. Across training objects, light and heavy demonstrations have fingertip force ranges of approximately $0.3$--$5.8$~N and $3.5$--$11.9$~N, respectively.
    \item \textbf{Rotation}: Success requires rotating an object to the target angle in the reference-specified direction. Transfer tests include screw turning and cube rotation. For cap rotation, we vary the reference direction to test whether the policy follows clockwise and counterclockwise demonstrations.
    \item \textbf{Insertion}: The robot establishes sliding contact with a receptacle and aligns the connector. Success requires reaching the target insertion depth. We evaluate on unseen USB, Type-C, and two-prong connectors to test adaptation to different connector geometries.
\end{itemize}

Together, these tasks cover complementary challenges in contact-rich manipulation: controlling the intensity of sustained contact, following specified rotation directions, and coordinating transitions between different contact states.

\subsection{One-Shot Transfer and Behavior Specification}

\subsubsection{One-Shot Transfer to Unseen Instances}

We first evaluate whether a single target-object demonstration enables a frozen policy to execute a contact-rich behavior on an unseen instance.
Table~\ref{tab:main_results} evaluates transfer to three unseen instances per task family. CTP achieves an overall success rate of $78.7\%$, compared with $65.7\%$ for BPP*, $40.7\%$ for DP-FT, and $37.0\%$ for $\pi_{0.5}$-FT. It improves over the strongest baseline, BPP*, by $13.0$ percentage points and achieves the best or tied-best performance on eight of the nine instances.

The gains over BPP* are most evident on wiping instances B and C and rotation instance C. Insertion remains more challenging: CTP improves on instances A and B but reaches $33.3\%$ on instance C, below BPP* at $41.7\%$. These results support one-shot transfer across the evaluated object instances with a frozen policy.



\begin{figure*}[!t]
    \centering
    \includegraphics[width=\textwidth,trim=0 4pt 0 0,clip]{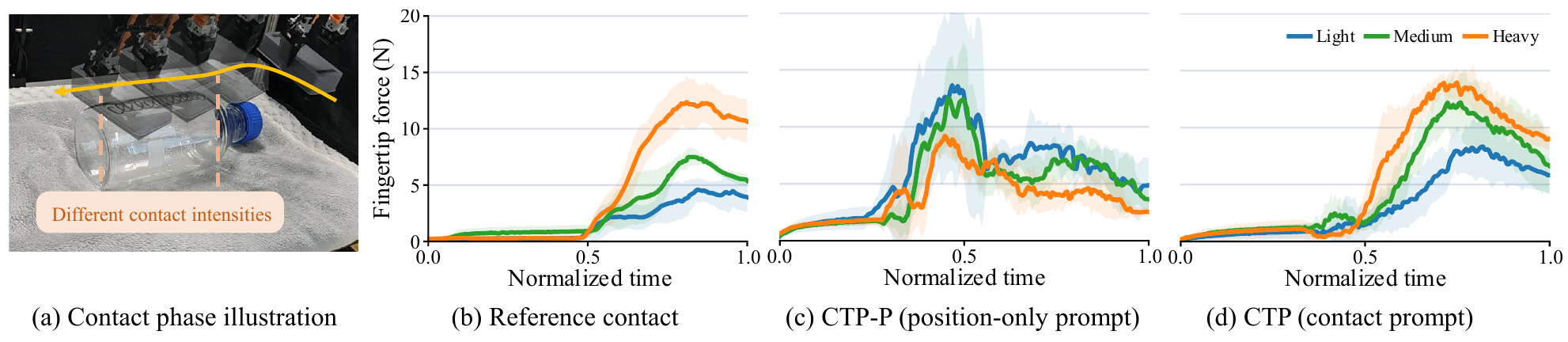}
    \vspace{-5mm}
    \caption{\textbf{Contact prompts specify graded wiping intensity.}
    Solid curves show mean fingertip force over normalized time; shaded regions indicate $\pm$1 standard deviation. Force is the fingertip force magnitude after subtracting the grasping baseline.
    (a)~Contact phase of wiping.
    (b)~Light, medium, and heavy reference trajectories.
    (c)~Executed force profiles with a position-only prompt (CTP-P).
    (d)~Executed force profiles with CTP.}
    \vspace{-2mm}
    \label{fig:reference_force}
\end{figure*}

\begin{figure}[!t]
    \centering
    \includegraphics[width=\columnwidth,trim=0 6pt 0 0,clip]{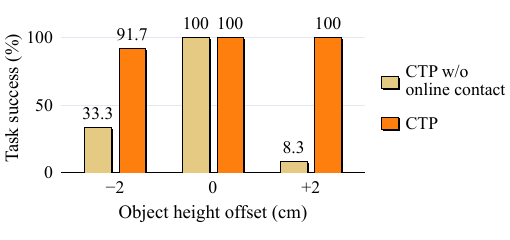}
    \vspace{-5mm}
    \caption{\textbf{Online contact grounds execution under height perturbations.} Task success rates of CTP and a variant without online contact observations at nominal height and under vertical shifts.}
    \vspace{-3mm}
    \label{fig:height_perturbation}
\end{figure}

\subsubsection{Contact Behavior Specification}

We evaluate whether changing only the reference trajectory induces corresponding behavior changes under a fixed task configuration. We consider wiping intensity and cap rotation direction, with mean behavior-conditioned scores reported in Fig.~\ref{fig:specification}(a).

For light and heavy wiping, CTP achieves scores of $91.7\%$ and $83.3\%$, respectively, compared with $75.0\%$ and $58.3\%$ for BPP*. For clockwise and counterclockwise cap rotation, CTP achieves $100.0\%$ in both conditions, compared with $91.7\%$ and $83.3\%$ for BPP*. Both methods follow the demonstrated rotation direction with high scores, while CTP shows a larger advantage in following the reference-specified wiping intensity. Together with the force profiles in Fig.~\ref{fig:reference_force}, these results support the use of contact trajectories to specify distinct interaction intensities even when the demonstrated motions are similar.

We also visualize the prompt embeddings using PCA in Fig.~\ref{fig:specification}(b), where each point represents one reference trajectory. The distributions vary with wiping intensity and rotation direction within each task, providing qualitative evidence of behavior-related variation in the prompt representations.

\subsection{Understanding Contact Trajectory Prompting}

CTP uses contact information in both the reference prompt and online observations. We ablate these two pathways separately to distinguish behavior specification from online physical grounding.

\subsubsection{Reference Contact Specifies Physical Interaction}
To isolate contact information as a \emph{behavior specification}, we compare CTP with CTP-P, which retains only positions in the reference prompt while keeping online contact inputs. Training demonstrations are collected in two wiping conditions, light and heavy, with no separate medium-intensity group. Both policies are trained on these demonstrations and evaluated on unseen objects using light, medium, and heavy references. The reference motions and timing are similar across conditions, while their force levels differ.

Figure~\ref{fig:reference_force} compares reference and executed force profiles. CTP-P produces overlapping profiles that fail to preserve the reference intensity ordering. CTP preserves the light--medium--heavy ordering, producing an intermediate force level in response to the medium reference. This comparison suggests that reference contact information enables graded intensity specification beyond selecting between the two training categories. The position-only prompt does not preserve this intensity distinction in our evaluation.

\begin{table}[t]
\centering
\caption{Ablation of prompt representation and injection.}
\label{tab:representation_ablation}
\small
\resizebox{0.88\columnwidth}{!}{%
\begin{tabular}{lcc}
\toprule
Variant & Nominal & Height perturbed \\
\midrule
\multicolumn{3}{l}{\textit{Prompt representation}} \\
Raw concat            & $77.8\%$ & $52.8\%$ \\
Encoder, from scratch & $88.9\%$ & $61.1\%$ \\
Pretrained encoder    & $\mathbf{97.2\%}$ & $\mathbf{94.4\%}$ \\
\midrule
\multicolumn{3}{l}{\textit{Prompt injection}} \\
Mean pooling          & $80.6\%$ & $38.9\%$ \\
Cross-attention       & $\mathbf{97.2\%}$ & $\mathbf{94.4\%}$ \\
\bottomrule
\end{tabular}%
}
\vspace{-3mm}
\end{table}

\subsubsection{Online Contact Grounds Closed-Loop Execution}
To evaluate \emph{online physical grounding}, we compare CTP with a variant that removes online contact inputs while retaining the contact trajectory prompt. We shift the test object vertically by $\pm2$~cm relative to the reference configuration, changing when contact occurs during execution.

As shown in Fig.~\ref{fig:height_perturbation}, CTP achieves $95.8\%$ success averaged over the two perturbed heights, compared with $20.8\%$ without online contact. With online contact observations, the policy continues the wiping stroke after earlier or later contact; without them, execution often fails to establish surface contact. Together with the reference-contact ablation, this result supports complementary roles: the reference specifies the desired interaction, and online contact observations ground its execution in the current physical state.

\subsection{Architecture Ablations}
Table~\ref{tab:representation_ablation} evaluates temporal contact representation and observation-conditioned prompt retrieval on wiping objects used during training. We report task success at nominal and perturbed heights to assess performance under changes in contact geometry.

\subsubsection{Temporal Contact Representation}
Replacing raw concatenation of end-effector positions, bilateral distributed force maps, and fingertip wrenches with a temporal prompt encoder trained from scratch raises success from $77.8\%$ to $88.9\%$ at nominal height and from $52.8\%$ to $61.1\%$ under height perturbations. Reconstruction pretraining further raises success to $97.2\%$ and $94.4\%$, respectively. The larger improvement under perturbations supports the value of reconstruction pretraining for robust execution under changes in contact geometry.

\subsubsection{Observation-Conditioned Prompt Retrieval}
Replacing query-to-prompt cross-attention with mean pooling and concatenation reduces success from $97.2\%$ to $80.6\%$ at nominal height and from $94.4\%$ to $38.9\%$ under perturbations. The larger drop under perturbations supports retaining ordered prompt tokens and retrieving reference information using online observations, particularly when the execution geometry differs from the reference.


\section{Conclusion}


We presented Contact Trajectory Prompting (CTP) for one-shot transfer of contact-rich behaviors. CTP encodes demonstrated motion and contact information into an ordered prompt that guides a frozen policy through online observations, enabling closed-loop execution without parameter updates. Real-robot experiments on wiping, rotation, and insertion demonstrate transfer to unseen object instances and control of contact intensity and manipulation direction by changing only the reference demonstration. Ablations support complementary roles: reference contact specifies the desired behavior, while online feedback grounds execution in the current contact state. These findings establish contact trajectories as effective behavioral specifications for in-context robot learning beyond motion imitation.

Building on these findings, future work will extend CTP to more diverse tasks, richer temporal interaction patterns, and different robotic platforms.



\bibliographystyle{IEEEtran}
\bibliography{refs/main}

\end{document}